\documentclass[sigconf, nonacm]{acmart}

\usepackage{tikz, pgfplots}
\usepackage{pgfplotstable,booktabs}

\pgfplotsset{compat=1.17}

\usetikzlibrary{external}
\usepackage{subfig}

\newcommand{\vect}[1]{\boldsymbol{#1}}

\newcolumntype{C}[1]{>{\centering\arraybackslash}m{#1}}

\AtBeginDocument{%
  \providecommand\BibTeX{{%
    \normalfont B\kern-0.5em{\scshape i\kern-0.25em b}\kern-0.8em\TeX}}}

\renewcommand\footnotetextcopyrightpermission[1]{} 

\begin{document}

\title{Multi-Head Online Learning for Delayed Feedback Modeling}

\author{Hui Gao}
\email{hui.gao@airbnb.com}
\affiliation{
	\institution{Airbnb}
	\streetaddress{888 Brannan Street}
	\city{San Francisco}
	\state{California}
	\postcode{94103}
}

\author{Yihan Yang}
\email{yihan.yang@bytedance.com}
\affiliation{
	\institution{Bytedance Inc.}
	\streetaddress{250 Bryant St}
	\city{Mountain View}
	\state{California}
	\postcode{94041}
}

\renewcommand{\shortauthors}{Gao and Yang}

\begin{abstract}
In online advertising,  it is highly important to predict the probability and the value of a conversion (e.g., a purchase). It not only impacts user experience by showing relevant ads, but also affects ROI of advertisers and revenue of marketplaces.  Unlike clicks, which often occur within minutes after impressions, conversions are expected to happen over a long period of time (e.g., 30 days for online shopping). It creates a challenge, as the true labels are only available after the long delays. Either inaccurate labels (partial conversions) are used, or models are trained on stale data (e.g., from 30 days ago). The problem is more eminent in online learning, which focuses on the live performance on the latest data. In this paper, a novel solution is presented to address this challenge using multi-head modeling. Unlike traditional methods, it directly quantizes conversions into multiple windows, such as day 1, day 2, day 3-7, and day 8-30. A sub-model is trained specifically on conversions within each window. Label freshness is maximally preserved in early models (e.g., day 1 and day 2), while late conversions are accurately utilized in models with longer delays (e.g., day 8-30).  It is shown to greatly exceed the performance of known methods in online learning experiments for both conversion rate (CVR) and value per click (VPC) predictions. Lastly, as a general method for delayed feedback modeling, it can be combined with any advanced ML techniques to further improve the performance.

\end{abstract}

\ccsdesc[500]{Computing methodologies~Modeling methodologies}
\ccsdesc[100]{Computing methodologies~Online learning settings}
\ccsdesc[100]{Computing methodologies~Learning linear models}

\keywords{VPC modeling, CVR modeling, conversion delays, delayed feedback}

\maketitle

\section{Introduction}
As a fast growing segment, global digital advertising spend exceeded \$378 billion in 2020. Among it, a large portion is spent in programmatic online auctions, such as search ads on Google and display ads on Facebook. Bids can be placed at multiple levels such as CPM (cost per thousand impressions), CPC (cost per click), and CPA (cost per action). Smart bidding algorithms like Google TROAS (Target Return On Ad Spend) can be used to optimize the target efficiency without explicitly specifying bid values. Regardless of the bidding algorithms, it is fundamental in online auctions to accurately predict conversions. As the ultimate goal of online advertising, it is of the interest of both advertiser and marketplaces.

Conversions are often modeled at the click level. For example, conversion rate (CVR) is defined as the probability of a conversion after a click. When it is combined with the expected value of a conversion (CV), the value per click (VPC) can be derived in Eq. \ref{eq_cvr_vpc}, assuming CV is independent of CVR.  Alternatively, VPC can be directly modeled for each click. Instead of logistic regression, a linear regression can be trained to directly fit the observed conversion values. 

\begin{equation} 
\label{eq_cvr_vpc}
	VPC = CVR * CV
\end{equation}

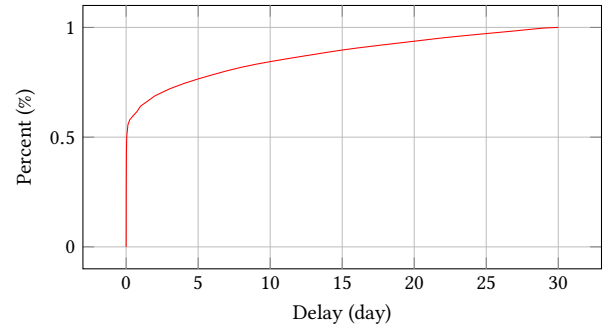
\begin{figure}
	\small
	\begin{tikzpicture}
		\begin{axis}[
			width=\columnwidth,
	         height=0.6\columnwidth,
	         grid=major,
			xlabel=Delay (day),
			xtick=\empty,
			extra x ticks={0,  7200,  14400, 21600, 28800, 36000, 43200},
			extra x tick labels={0, 5,10,15, 20, 25, 30},
			ylabel=Percent (\%),
			xlabel near ticks,		
			ylabel near ticks,
		]
			\addplot[color=red] table[col sep=comma] {charts/conversions.csv};
		\end{axis}
	\end{tikzpicture}
    \caption{Percentage of Total Conversions by Delay}
    \label{fig_conversion_delays}    
\end{figure}

While there are trade-offs between the two approaches, both are commonly used in conversion modeling.  However, they face the same challenge of long conversion delays. For example, in online shopping applications, a conversion can happen any time within 30 days after a click. As shown in Fig. \ref{fig_conversion_delays} for the benchmark 2020 Criteo dataset used in the experiment, the first day alone captured over 58.7\% of conversions. The remaining conversions spread evenly over the long period of time. Obviously, waiting for the true labels after 30 days ({\em shifted} model) will cause severe staleness in the prediction, especially in online applications. Feeding clicks as negative examples and conversions as positive examples ({\em naive} model), whenever they occur, will result in large biases due to Positive Unlabeled and Missing-Not-At-Random problems \cite{Saito2020}.  

There are lots of effort to address the challenge of delayed feedback modeling. Either the loss functions are modified to include delay terms (e.g., \cite{Chapelle2014, Saito2020, Yang2021}) or the importance weights on examples are predicted to correct the prediction bias (e.g., \cite{Ktena2019, Yasui2020, Gu2021}). While they offer significant improvements over the baseline (e.g., {\em shifted} model), gaps remain to the model trained on oracle labels, which are artificially extracted on historical data assuming no delays.

The problem is more challenging for online learning \cite{Hoi2021}, which often is a preferred technique for online applications due to its focus on the accuracy for the sequence of predictions (e.g., at the moment, not the average CVR or VPC over the 3 month training window). Since \cite{Chapelle2014, Saito2020} are built on modeling the elapsed time between clicks and conversions, they are not suitable in online learning or continuous training, unless extensions like \cite{Yang2021} are enabled. On the other hand, although the importance weight based methods can work in online learning, there are clear limitations. For example, the approach in \cite{Yasui2020} requires a counter-factual window of 5 days, which is often too long for online applications. Other methods like \cite{Ktena2019, Gu2021} are designed to have no or a very small wait window (e.g., 15 mins). But they are trained on the biased input with Unlabeled Positive and Missing-Not-At-Random problems. They rely on importance weights derived from the model prediction and additional examples to correct the bias. However, they may not be stable due to the positive feedback loop. And their convergence is not guaranteed, given the stochastic nature in SGD training. 

In this paper, a novel solution based on multi-head modeling is proposed to explicitly address the challenge. The conversions are quantized based on the delay of their occurrence. (e.g., day 1, day 2, day 3-7, and day 8-30). A sub-model is specifically trained for conversions in each time range, using the accurate labels without any bias. Since the recent sub-models, such as day 1 and day 2, can be trained with no or little delays, model freshness is maximally preserved for online learning. Furthermore, only a small portion of total conversions (e.g., only 20.0\% of conversion happened after 7 days as shown in Fig. \ref{fig_conversion_delays}) are covered by the latter sub-models. Despite their longer delays (e.g., 30 days for the last model), they become much less a concern, especially when considering the benefit of improved label accuracy. As the result, the combined output of sub-models significantly out-performed the existing methods, getting much closer to the performance of the {\em oracle} model for both CVR and VPC predictions. 

Furthermore, the proposed multi-head modeling is perfectly suited for online learning. Multiple sub-models can be updated incrementally using the standard online learning approach. It is flexible to handle any conversion delays, as high as 1 year or as short as a few minutes, by properly designing quantization windows. It is also model agnostic, which can be combined with any ML modeling techniques like deep neural networks to further improve the modeling accuracy. It is a significant breakthrough in delayed feedback modeling, despite it being a simple and elegant solution. 

\section{Related Work}
Conversion modeling have been extensively studied in the literature.  Recent work includes \cite{Rosales2012, Chapelle2015, Juan2017} for CVR modeling and \cite{Sodomka2013, Yuan2018} for VPC modeling. While various models and techniques have been proposed, it is common to deploy Logistic Regression for CVR prediction and Linear Regression for VPC prediction. 

The challenge of long conversion delays was mainly studied in the application of CVR prediction. It was first examined in \cite{Chapelle2014} using Delay Feedback Model (DFM). By assuming the positive delays follow the Exponential Distribution, it jointly optimized the loss function using either the expectation-maximization (EM) \cite{Mclachlan2007} or stochastic gradient descent algorithm \cite{Mitchell1997}. It was shown to exceed the best baseline model ({\em shifted}) on the public 2013 Criteo dataset, while the gap to the {\em oracle} model remains large.

Later work \cite{Safari2017, Tallis2018, Yoshikawa2018, Ktena2019} further expands the research on conversion delays by examining various methods to speed up the model training, mix short-term and long-term signals, estimate delays for non-exponential distributions, and calibrate negative examples. Although \cite{Ktena2019} demonstrated significant performance improvements on the Twitter private data set, DFM still won (not statistically significant) on the public 2013 Criteo dataset. Following the path of DFM, one recent effort in \cite{Saito2020} achieved similar performance using inverse propensity score (IPS) to perform unbiased CVR Prediction. It can be used in applications where the delay does not follow exponential distribution. Lastly, DFM was extended to online learning in \cite{Yang2021} using elapsed time sampling. 

The recent approach FSIW \cite{Yasui2020} is the first technique shown to out-perform DFM on the public 2013 Criteo dataset.  Its labels for the CVR model were collected in the 5-day counterfactual window. To correct the partial conversions observed in the small training window, 2 separate models were trained to predict importance weights for positive and negative labels. It was shown to exceed the performance of DFM, while reducing the gap to the {\em oracle} model considerably. 

The latest DEFER model \cite{Gu2021} extends the importance weighting algorithm in \cite{Ktena2019} to include real negatives. In addition to correcting for fake negatives, it duplicates the full dataset after the conversion window (e.g. 30 days) to train a deep neural networks. The bias was corrected by using importance weights derived from the existing CVR predictor and a fake negative predictor. Although DEFER can work in online learning in real time, the optimal performance was achieved using a wait window of 15 mins, which captures 35\% of conversions in the public 2013 Criteo dataset and 55\% in the private TaoBao dataset. This highlights the importance of accurate labels to reduce modeling bias. 

Although DEFER is approaching the problem differently from FSIW, their mathematical formula are highly similar. Both utilizes importance weights to handle delayed feedback. In the DEFER model, the importance weights are computed using the existing CVR predictor (self-dependent and biased) and a false negative predictor. On the other hand, the FSIW model trains the two importance weight models independently, which is more stable and accurate. However, there is one major difference of the DEFER model. It feeds the duplicated data after the full conversion window (e.g., 30 days), which effectively utilizes the conversions after the initial wait window, while the FSIW model never feeds any positive examples outside the counter-factual window. It could result in much improve performance on bias reduction. However,there is no direct comparison of DEFER to FSIW in \cite{Gu2021}, despite of the reference. 

This paper is significant in that it proposes a novel approach of Multi-Head Online Learning (MHOL) to model delayed feedback of conversions. Despite its simplicity, it is shown in Sect. \ref{sect_experiment} to out-perform the latest methods of DEFER and FSIW in CVR prediction using logistic regression on the 2020 public Criteo dataset. Although similar approaches of delay quantization may have been studied (e.g., \cite{Badanidiyuru2021}), no prior research was known to rigorously benchmark the method in the context of delayed feedback modeling on the public standard dataset. Moreover, this approach is model agnostic by directly quantizing the labels. As demonstrated on VPC prediction using linear regression, it can be combined with any advanced ML techniques to further improve the accuracy of delayed feedback modeling.

\section{Multi-Head Online Learning}
\begin{figure}
	\includegraphics[width=\linewidth]{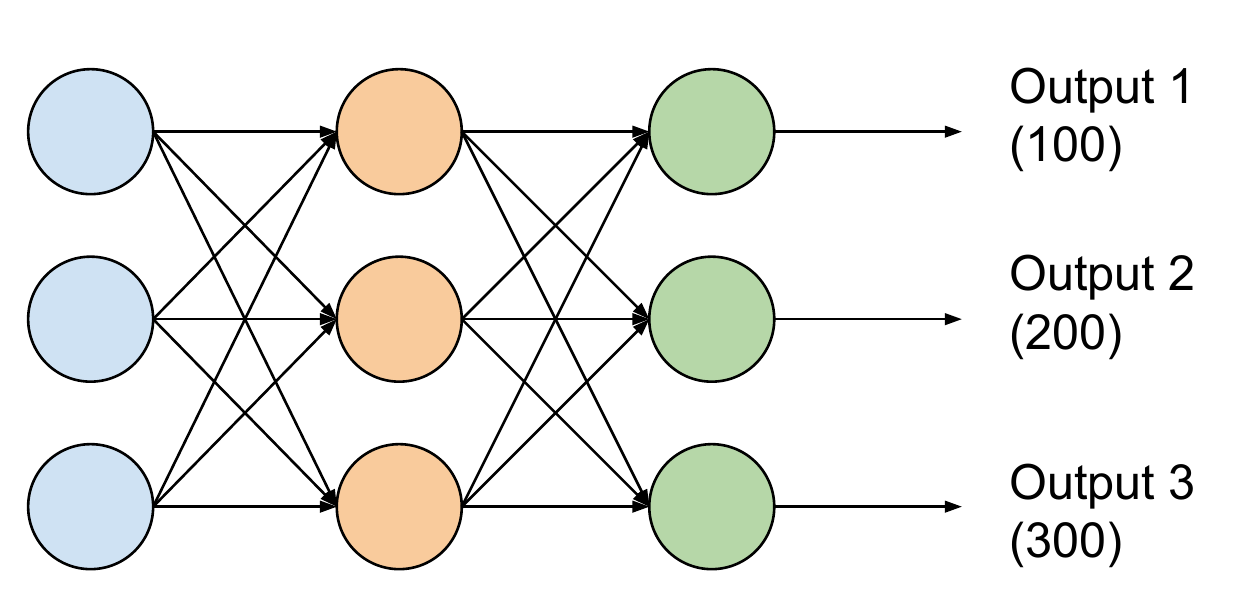}
	\caption{Multi-Head Modeling in Neural Network}
	\label{fig_multi_head} 
\end{figure}

Multi-head modeling is commonly used technique in neural networks (see illustration in Fig. \ref{fig_multi_head}) including deep learning. Different models share the input features or the hidden layers, but predict different outputs based on the different training labels. Applications, such as multi-head attention \cite{Vaswani2017} in NLP, was shown to be a breakthrough in machine translation. This technique also fits perfectly to address the challenge of long conversion delays, once the labels are quantized into multiple conversion windows, as presented in the following sections. 

\subsection{Label Quantization}
In the domain of conversion prediction, a input feature vector $X$ is extracted for every click. Labels are defined as either a binary variable $Y$ of whether a conversion occurred in the full observation window $r$ for CVR prediction or a real variable $V$ for the actual value of conversion for VPC prediction.  The mathematical definitions are outlined below. 

\begin{itemize}
  \item $X$: $\mathbb{X}$ valued random variable of features for each click;
  \item $r$: a time range constant (e.g., the maximum wait time) to observe conversions after clicks. It is defined based on application, such as 30 days for online shopping;
  \item $Y$: $\{0, 1\}$-valued random variable indicating whether a conversion occurred within $r$;
  \item $V$: $\mathbb{R}+$ valued random variable representing the value of a conversion within $r$; 
  \item $D$: $\mathbb{R}+$ valued random variable of the delay between a click and its conversion. It is undefined when there are no conversions within $r$. 
\end{itemize}

The challenge of long conversion delays is due to the fact that $Y$ and $V$ are available only after the full wait window of $r$. Models trained on the delayed labels are stale and less accurate to reflect the current trend. To address this issue, the full conversion window $r$ can be quantized into smaller windows $r_i$. For example, if the maximum wait time is 30 days, it can be split into 30 windows, one for each day of delay. Conversions in each day can be modeled by one sub-model independently. The combined output of sub-models can be used to predict the full conversions within $r$. Mathematically, it is defined as the following. 

\begin{itemize}
  \item $t_i$: $\mathbb{R}+$ valued constant sequences of time after clicks.  $t_0=0$,  $t_0 < t_1 < t_2 < ... < t_n$, and $r=(t_0, t_n]$;
  \item $r_i$: a time range of $(t_{i-1}, t_i]$ to observe conversions;
  \item $Y_i$: $\{0, 1\}$-valued random variable indicating whether a conversion occurred within time range $r_i$. If $d \in r_i$, $y_i=y$. Otherwise, $y_i=0$;
  \item $V_i$: $\mathbb{R}+$ valued random variable representing the value of a conversion within time range $r_i$. If $d \in r_i$, $v_i=v$. Otherwise, $v_i=0$;   
\end{itemize}

Because $r_i$ fully represents $r$ as quantized ranges (Eq. \ref{eq_sub_model_r}), $Y$ and $V$ can be represented using $Y_i$ and $V_i$ as in Eq. \ref{eq_sub_model_y} and Eq. \ref{eq_sub_model_v} respectively, assuming only a single conversion if it occurs.

\begin{equation} 
\label{eq_sub_model_r}
	r = \sum_{i=1}^{n} r_i
\end{equation}

\begin{equation} 
\label{eq_sub_model_y}
	Y = \sum_{i=1}^{n} Y_i
\end{equation}

\begin{equation} 
\label{eq_sub_model_v}
	V = \sum_{i=1}^{n} V_i
\end{equation}

\subsection{Multi-Head Modeling}
Given multiple labels of $Y_i$ and $V_i$, multi-head modeling can be applied for both CVR and VPC predictions. Assuming Logistic Regression is used to model CVR and Linear Regression is used to model VPC, the models can be trained by minimizing negative cross entropy (Eq. \ref{eq_neg_logloss}) and sum squared error (Eq. \ref{eq_sse}) respectively.  The weight vector $\vect{w_c}$ and $\vect{w_v}$ will be used in Eq. \ref{eq_cvr} and Eq. \ref{eq_vpc} to predict $p(\vect{x})$ as CVR and $\hat{v}(\vect{x})$ as VPC.

\begin{equation} 
\label{eq_cvr}
	\hat{y} = p(\vect{x}) = \frac{1}{1 + exp(-\vect{w} \cdot \vect{x})}
\end{equation}

\begin{equation} 
\label{eq_neg_logloss}
	\vect{w_{c}} = \underset{\vect{w}}{argmin}- \sum{(y \cdot log(p(\vect{x})) + (1-y) \cdot log(1-p(\vect{x})))}
\end{equation}

\begin{equation} 
\label{eq_vpc}
	\hat{v}(\vect{x}) = \vect{w} \cdot \vect{x}
\end{equation}

\begin{equation} 
\label{eq_sse}
	\vect{w_{v}} = \underset{\vect{w}}{argmin}\sum{(v - \hat{v}(\vect{x}))^2}
\end{equation}

By using multiple labels of $Y_i$ and $V_i$, Eq. \ref{eq_neg_logloss} and Eq. \ref{eq_sse} can be extended to train multiple models based on the same set of input examples as in Eq. \ref{eq_neg_logloss_mh} and Eq. \ref{eq_sse_mh} respectively for $i$ in $[1, n]$. Each represents one output of the multi-head modeling. 

\begin{equation} 
\label{eq_neg_logloss_mh}
	\vect{w_{c, i}} = \underset{\vect{w}}{argmin}- \sum{(y_i \cdot log(p(\vect{x})) + (1-y_i) \cdot log(1-p(\vect{x})))}
\end{equation}

\begin{equation} 
\label{eq_sse_mh}
	\vect{w_{v, i}} = \underset{\vect{w}}{argmin}\sum{(v_i - \hat{v}(\vect{x}))^2}
\end{equation}

To predict the overall CVR and VPC, Eq. \ref{eq_cvr_mh} and Eq. \ref{eq_vpc_mh} can be used to simply combine sub-model outputs, based on Eq.\ref{eq_sub_model_y}, Eq.\ref{eq_sub_model_v}, Eq. \ref{eq_cvr}, and Eq. \ref{eq_vpc}.

\begin{equation} 
\label{eq_cvr_mh}
	\hat{y} = p(\vect{x}) = \sum_{i=1}^n \frac{1}{1 + exp(-\vect{w_{c,i}} \cdot \vect{x})}
\end{equation}

\begin{equation} 
\label{eq_vpc_mh}
	\hat{v}(\vect{x}) = \sum_{i=1}^n(\vect{w_{c,i}} \cdot \vect{x}) = \sum_{i=1}^n{\vect{w_{c,i}}} \cdot \vect{x}
\end{equation}

Since label quantization can be utilized for any models, the proposed multi-head modeling approach is model agnostic. In addition to Logistic Regression and Linear Regression, it can be combined with any model in conversion prediction, such as multi-layer neural networks with embedding. Unlike the classic DFM model \cite{Chapelle2014}, there is no modification to the loss functions or assumption of exponential delays. It is a general technique to model delayed feedback.

The conversion delays are handled explicitly by quantizing labels into multiple time ranges. Both $Y_i$ and $V_i$ are only available at the time $t_i$ after the click. For example, the earliest labels $Y_1$ and $V_1$ are available after $t_1$, which can be as short as 15 mins based on the distribution of conversion delays in the application. The last labels $Y_n$ and $V_n$ are available after $t_n$, which is the full conversion window. As shown in Fig. \ref{fig_conversion_delays}, majority of conversions (58.7\%) happened in the first day of clicks in the public 2020 Criteo dataset. It is beneficial to have smaller, but fresher early labels, whereas having longer ranges to later labels. This enables multi-head modeling to react to majority of signals with minimal delays. 

At the same time, the computational complexity of the multi-head modeling is proportional to the number of sub-models. It is desirable to use less ranges to bucket the labels. Furthermore, if the range is extremely narrow, the models may not have enough positive labels to generalize. For example, if there is a sub-model for every 1s after clicks or every 5 mins after 20 days, many sub-models will have very few positive labels. Their convergence and stability are at risk. Lastly, it is time consuming to tune hyper parameters for multi-head models. If it takes 3 days to sweep 2 hyper parameters for one model, it could take 300 days for 100 sub-models. And inconsistent or over-tuning may result in the performance degradation in the combined output. 

Therefore, it is recommended to quantize the labels so that the similar number of positive examples occur in each range. For example, if there are 5 sub-models, it is ideal to have each of them capturing 20\% of conversion volumes. This can be practically determined by analyzing the distribution of conversion delays (e.g., building the histogram of conversion delays). It will not only allow as many as possible early labels, but also reduce the overall number of sub-models by having fewer, but wider range labels with longer delays. This also ensures all sub-models to share the same hyper parameters, since they are trained on the same feature set and similar positive label ratios.  

\subsection{Online Learning}
The multi-head modeling can seamlessly support online learning. This can be done on real-time click stream, as well as a fixed batch size, such as daily online learning. 

\begin{figure}
	\includegraphics[width=\linewidth]{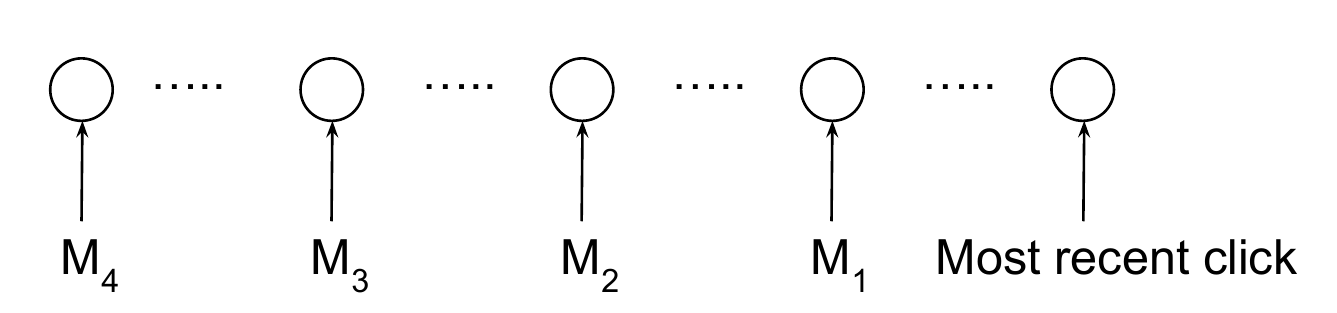}
	\caption{Multi-Head Online Learning on Real-Time Click Stream. Sub-models ($M_1$ to $M_4$) wait until different delays ($t_1$ to $t_4$) after their corresponding click time. Label is determined as whether a conversion happened in $r_i = (t_{i-1}, t_i]$ after the click.}
	\label{fig_multi_head_online_learning} 
\end{figure}

As illustrated in Fig. \ref{fig_multi_head_online_learning}, sub-models $M_1$ to $M_4$ are training at different positions in the click stream. They only consume the click after waiting $t_i$. The labels are decided based on the discretized range $r_i = (t_{i-1}, t_i]$. If a conversion happened within $r_i$, the click is labeled as positive (1 for CVR or conversion value for VPC). Otherwise, it is labeled as negative or 0. Each model $M_i$ only advances to the next click in the stream, after it has waited $t_i$ after the click. Assuming $t_1$ is 5 mins,$t_2$ is 1 hour, $t_3$ is 1 day, and $t_4$ is full 30 days, the proposed multi-head models will train $M_1$ with 5 mins delay to predict conversion in $(0, t_1]$. Similarly, model $M_2$ has a delay of 1 hour and predicts conversions in $(t_1, t_2]$, and etc. By adding the output of all sub-models, conversions can modeled without any labeling bias and minimal delays. The architecture is also computationally efficient as each sub-model only needs to maintain a single pointer in the click stream (e.g., Kafka queue). Conversions can be stored using memcache on click\_id to ensure fast lookup and efficient update, when they are received.

Alternatively, multi-head online learning can be configured at a fixed batch size (partition). For example, certain applications only require daily update like bidding or data is not available in real-time. In this case, the quantization ranges need to be aligned to the partitions. Conversions may overflow into one additional batch. For example, for a 30 day conversion window, clicks may happen at the end of day 1. Their conversions may happen within 30 days after day 1, which may be in the 31th daily partition. Hence one additional input is required in online batch training.

\section{Experiment}
\label{sect_experiment}
To effectively evaluate the performance of the proposed method, the latest 2020 Criteo\footnotemark \footnotetext{https://ailab.criteo.com/criteo-sponsored-search-conversion-log-dataset/} dataset is used.  It contains logs from Criteo Predictive Search from Aug. 3, 2020 to Oct. 16, 2020. The widely referenced 2013 Criteo\footnotemark \footnotetext{https://labs.criteo.com/2013/12/conversion-logs-dataset/} dataset is no longer publicly accessible.

Despite a different date range, the new 2020 Criteo dataset is similar to the 2013 dataset.  Each record represents a click on a product related advertisement.  Features include product information (e.g., age, brand, gender, price), time of the click, user characteristics, and device information. All categorical features have been hashed to protect privacy. Conversion is defined as whether the product was bought within a 30 day window after the click. Purchase value as well as time delay of conversion is reported. 

The proposed Multi-Head Online Learning (MHOL) model is configured at the daily interval of online learning. There are 5 sub-models defined at day 1, day 2, day 3-5, day 6-12, and day 13-31. Each sub-model cover 64.1\%, 4.7\%, 7.7\%, 10.1\%, and 13.4\% of total conversions respectively. 

For the ease of comparison, the experiment setup is identical to \cite{Chapelle2014,Yasui2020}. Features are pair-wise crossed and hashed to 24 bits. The average performance over the last 7 days are used for testing. For CVR prediction, the logistic regression model is trained with the Negative Logloss (NLL) as the primary metric. Relative Cross Entropy (RCE) is computed to simplify the comparison.  For VPC prediction, the linear regression model is trained with the Mean Squared error (MSE) as the primary metric. The average bias is used as the secondary metric to measure the convergence. 

Training is performed in TensorFlow 2.6 \cite{tensorflow2015} using Adam \cite{Kingma2015} optimization and L2 regularization. Parameters are extensively tuned to report the best performance of each model. The following models are benchmarked, reflecting the state of the art. The FSIW model \cite{Yasui2020} is selected for its superior performance to DFM \cite{Chapelle2014}. The DEFER model \cite{Gu2021} is selected for exceeding the performance of previous models \cite{Ktena2019, Yang2021}.

\begin{itemize}
\item{{\em Shifted}: true labels used after 30 days}
\item{{\em Oracle}: oracle labels used without any delays}
\item{{\em FSIW}: 5 day delayed partial labels with correction \cite{Yasui2020}}
\item{{\em DEFER}: real negative correction with 1 day wait window \cite{Gu2021}}
\item{{\em MHOL}: the proposed multi-head online learning model}
\end{itemize}

As shown in Table. \ref{table_cvr_prediction}, the proposed MHOL model achieved the best RCE at $14.726$ in CVR prediction. It significantly reduced the gap to the {\em oracle} model to $1.077$. The DEFER model was the 2nd best, narrowly beating FSIW by $0.1$ RCE. However, it showed large variances in the experiments. Despite the best record at $14.307$, its RCE can drop as low as $13.362$ at the same parameter. Although model warm-ups may help, the algorithm is by design unstable due to its self-feedback loop. After a long run, it may deviate significantly from the ideal state. Lastly, its importance weights formula can results in unexpected values, especially when denominators are close to $0$. A range bound has to be added to safe-guard the importance weights. In comparison, other models showed small RCE variations (about $0.01$) across multiple runs. The proposed MHOL model was clearly demonstrated to be the best algorithm for delayed feedback modeling in CVR prediction.

\begin{table}
	\begin{filecontents*}{charts/cvr_prediction.csv}
,NLL,RCE, Diff
{\em Shifted},0.287625,12.983946,2.818178
FSIW,0.283579,14.208054,1.59407
DEFER,0.283253,14.306626,1.495498
\textbf{MHOL},0.281913,14.725577,1.076547
{\em Oracle},0.27831,15.802124,
\end{filecontents*}
\pgfplotstabletypeset[		
		col sep = comma,
		every head row/.style={before row=\toprule, after row=\midrule},
		every last row/.style={before row=\midrule,, after row=\bottomrule},
		every row no 0/.style={after row=\midrule},		
		display columns/0/.style={string type, column name=, column type={m{6em}}},
		display columns/1/.style={zerofill, fixed, precision=4, column type={@{}C{6em}}},
		display columns/2/.style={zerofill, fixed, precision=3, column type={@{}C{6em}}},
		display columns/3/.style={zerofill, fixed, precision=3, column type={@{}C{6em}}},
	]
	{charts/cvr_prediction.csv}	
	\caption{CVR Prediction. Diff is reported as the gap to the RCE of the {\em Oracle} model. MHOL is highlighted as the best performing model.}
	\label{table_cvr_prediction}
\end{table}

Similarly in Table. \ref{table_vpc_prediction}, the proposed MHOL model achieved the lowest MSE at $10574.410$ in VPC prediction. The DEFER model delivered the 2nd best performance, which is close to FSIW. It showed similar stability issue in experiments. Notably, FSIW achieved the lowest bias (close to the {\em oracle} model) among all the candidate models. It is probably due to the 5 day counter-factual window, which is much recent to reflect the current trend. The bias of the proposed MHOL model was mainly coming of later sub-models (e.g., day 6-12, and day 13-31). The rest of the sub-models showed biases consistent with the {\em oracle} model. Nevertheless, it is still much lower than the {\em shifted} baseline model, which showed the highest bias in the experiment. This again demonstrated the effectiveness of the proposed MHOL model in VPC prediction for delayed feedback modeling.

To focus on the general technique of delayed feedback modeling, the DEFER model was trained using  the much simpler Logistic Regression and Linear Regression in daily online learning application. It is different than model in \cite{Gu2021}, which utilizes embedding and 15 mins wait window. It will be interesting in the future to evaluate the MHOL model in the real-time streaming applications using the deep neural networks. For model warm-up, a simple prior model (set as initial bias) was implemented to bootstrap FSIW and DEFER. Since the training dataset is sufficiently large (95 days), it is expected to result in minimal gaps to the pre-trained model in \cite{Gu2021}. 

\section{Conclusion}
In this paper,  a novel approach for delayed feedback modeling is proposed using multi-head online learning. By quantizing the conversions into multiple windows (e.g., day 1, day 2, day 3-7, and day 8-30), a sub-model is trained on conversions in each time range. It retains the maximum label freshness in the recent models, while ensures the label accuracy in the later models. The combined output was shown to significantly out-perform the existing methods, in both CVR (conversion rate) and VPC (value per click) predictions. Lastly, as a general method of delayed feedback modeling, multi-head online learning (MHOL) can be combined with any advanced ML modeling techniques, including multi-layer neural networks. It is capable to achieve 5-10 mins model update latency in online learning. This simple and elegant method is a break-through in conversion modeling with delayed feedback for online applications. 

\begin{table}
	\begin{filecontents*}{charts/vpc_prediction.csv}
,Bias,MSE,Diff
{\em Shifted},20.414,10616.51755,88.35955
FSIW,3.7419,10585.7559,57.5979
DEFER,14.4178,10583.06653,54.908528
\textbf{MHOL},16.8356,10574.4095,46.2515
{\em Oracle},3.2101,10528.1580,
\end{filecontents*}
\pgfplotstabletypeset[		
		col sep = comma,
		every head row/.style={before row=\toprule,after row=\midrule},
		every last row/.style={before row=\midrule,, after row=\bottomrule},
		every row no 0/.style={after row=\midrule},		
		display columns/0/.style={string type, column name=, column type={m{6em}}},
		display columns/1/.style={column name=Bias(\%), zerofill, fixed, precision=1, column type={@{}C{6em}}},
		display columns/2/.style={column name=MSE, zerofill, fixed, precision=3, column type={@{}C{6em}}},
		display columns/3/.style={column name=Diff, zerofill, fixed, precision=3, column type={@{}C{6em}}},
	]
	{charts/vpc_prediction.csv}	
	\caption{VPC Prediction. Diff is reported as the gap to the MSE of the {\em Oracle} model. MHOL is highlighted as the best performing model.}
	\label{table_vpc_prediction}
\end{table}

\bibliographystyle{ACM-Reference-Format}
\bibliography{references}

\end{document}